\documentclass{article}

\usepackage{times}
\usepackage{graphicx} % more modern
\usepackage{subfigure}

\usepackage{natbib}

\usepackage{algorithm}
\usepackage{algorithmic}

\usepackage{hyperref}

\usepackage{array}
\usepackage{booktabs}
\usepackage[table]{xcolor}% http://ctan.org/pkg/xcolor
\usepackage{amsmath}
\usepackage{paralist}

\usepackage{graphicx}
\usepackage{placeins}

\usepackage[arxiv]{icml2016}

\newcommand{\fig}[1]{Fig.~\ref{fig:#1}}
\newcommand{\tab}[1]{Table~\ref{tab:#1}}
\newcommand{\secc}[1]{Section~\ref{sec:#1}}

\icmltitlerunning{Learning Physical Intuition of Block Towers by Example}

\begin{document}

\twocolumn[
% \icmltitle{Learning Physical Intuition with Falling Blocks}
\icmltitle{Learning Physical Intuition of Block Towers by Example}

\icmlauthor{Adam Lerer}{alerer@fb.com}
\icmladdress{Facebook AI Research}
\icmlauthor{Sam Gross}{sgross@fb.com}
\icmladdress{Facebook AI Research}
\icmlauthor{Rob Fergus}{robfergus@fb.com}
\icmladdress{Facebook AI Research}

\icmlkeywords{physics, simulation, machine learning, deep learning}

\vskip 0.3in
]

\begin{abstract}
  Wooden blocks are a common toy for infants, allowing them to develop
  motor skills and gain intuition about the physical behavior of the
  world. In this paper, we explore the ability of deep feed-forward
  models to learn such intuitive physics. Using a 3D game engine, we
  create small towers of wooden blocks whose stability is randomized
  and render them collapsing (or remaining upright). This data allows
  us to train large convolutional network models which can accurately
  predict the outcome, as well as estimating the block
  trajectories. The models are also able to generalize in two
  important ways: (i) to new physical scenarios, e.g. towers with an
  additional block and (ii) to images of real wooden blocks, where it
  obtains a performance comparable to human subjects.
\end{abstract}

\section{Introduction}
\label{sec:intro}

Interaction with the world requires a common-sense understanding of
how it operates at a physical level. For example, we can quickly
assess if we can walk over a surface without falling, or how an object
will behave if we push it. Making such judgements does not require
us to invoke Newton's laws of mechanics -- instead we rely on intuition,
built up through interaction with the world.

In this paper, we explore if a deep neural network can capture this
type of knowledge. While DNNs have shown remarkable success on
perceptual tasks such as visual recognition \cite{alex12} and speech
understanding \cite{hinton2012deep},
they have been rarely applied to problems involving higher-level
reasoning, particularly those involving physical
understanding. However, this is needed to move beyond object classification
and detection to a true understanding of the
environment, e.g. ``What will happen next in this scene?''
Indeed, the fact that humans develop
such physical intuition at an early age \cite{Carey09}, well before most
other types of high-level reasoning, suggests its importance in
comprehending the world.

To learn
this common-sense understanding, a model needs a way to interact
with the physical world. A robotic platform is one option that has
been explored e.g.~\cite{Agrawal15}, but inherent complexities limit the diversity and
quantity of data that can be acquired.  Instead, we use Unreal Engine
4 (UE4) \cite{unreal}, a platform for modern 3D game development, to provide a
realistic environment. We chose UE4 for its realistic physics
simulation, modern 3D rendering, and open source license. We integrate the
Torch~\cite{Collobert11} machine learning framework directly into the UE4 game loop,
allowing for online interaction with the UE4 world.

\begin{figure}[t!]
\vskip 0.2in
\begin{center}
\centerline{\includegraphics[width=\columnwidth]{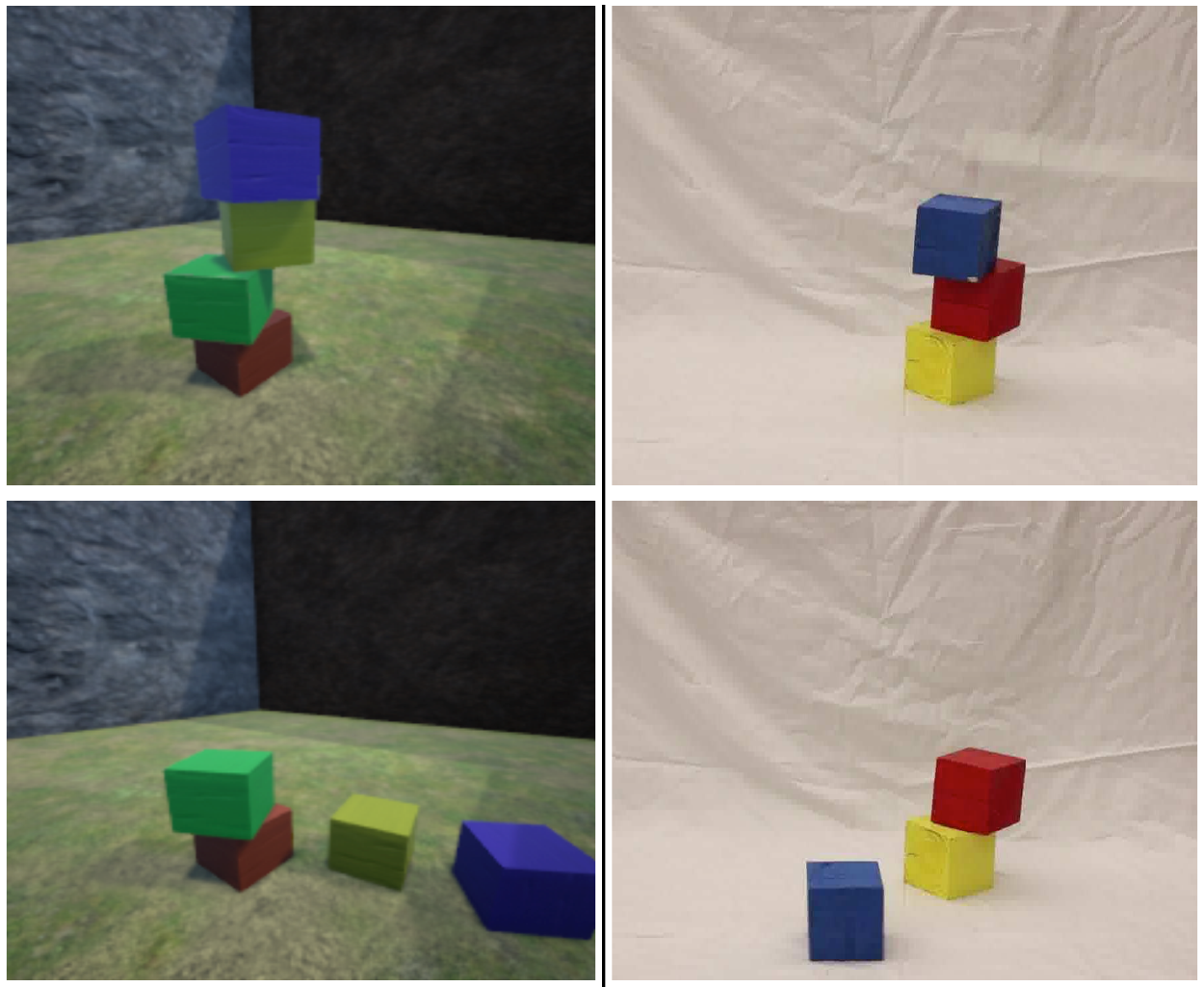}}
\caption{Block tower examples from the synthetic (left) and real (right) datasets.
  The top and bottom rows show the first and last frames
  respectively.}
\label{fig:teaser}
\end{center}
\vskip -0.2in
\end{figure}

One of the first toys encountered by infants, wooden blocks provide a
simple setting for the implicit exploration of basic Newtonian concepts such as
center-of-mass, stability and momentum. By asking deep models to
predict the behavior of the blocks, we hope that they too might internalize
such notions. Another reason for
selecting this scenario is that it is possible to construct real world
examples, enabling the generalization ability of our models to be
probed (see \fig{teaser}).

Two tasks are explored: (i) will the blocks fall over or not? and (ii)
where will the blocks end up? The former is a binary classification
problem, based on the stability of the block configuration. For
the latter we predict image masks that show the location of each
block. In contrast to the first task, this requires the models to
capture the dynamics of the system. Both tasks require an effective
visual system to analyze the configuration of blocks. We explore
models based on contemporary convolutional networks
architectures \cite{Lecun89}, notably Googlenet \cite{Ioffe15}, DeepMask \cite{Pinheiro15}
and ResNets \cite{He15}. While designed for classification or segmentation,
we adapt them to our novel task, using an integrated approach where the lower layers
perceive the arrangement of blocks and the upper layers implicitly capture
their inherent physics.

Our paper makes the following contributions:

{\bf Convnet-based Prediction of Static Stability:} We show
  that standard convnet models, refined on synthetic data, can
  accurately predict the stability of stacks of
  blocks. Crucially, these models successfully generalize to (i)
  new images of real-world blocks and (ii) new physical
  scenarios, not encountered during training. These models are purely
  bottom-up in nature, in contrast to existing approaches which rely
  on complex top-down graphics engines.

{\bf Prediction of Dynamics:} The models are also able to
  predict with reasonably accuracy the trajectories of the blocks as
  they fall, showing that they capture notions of acceleration and
  momentum, again in a purely feed-forward manner.

{\bf Comparison to Human Subjects:} Evaluation of the test
  datasets by participants shows that our models match their
  performance on held-out real data (and are significantly better on
  synthetic data). Furthermore, the model predictions have a
  reasonably high correlation with human judgements.

{\bf UETorch:} We introduce an open-source combination of the
  Unreal game engine and the Torch deep learning environment, that is
  simple and efficient to use. UETorch is a viable environment for a variety
  of machine learning experiments in vision, physical reasoning, and
  embodied learning.

\subsection{Related Work}

The most closely related work to ours is \cite{Battaglia13} who
explore the physics involved with falling blocks. A generative
simulation model is used to predict the outcome of a variety of block
configurations with varying physical properties, and is found to
closely match human judgment. This work
complements ours in that it uses a top-down approach, based on a
sophisticated graphics engine which incorporates explicit prior
knowledge about Newtonian mechanics. In contrast, our model is purely
bottom-up, estimating stability directly from image pixels and is
learnt from examples.

Our pairing of top-down rendering engines for data
generation with high capacity feed-forward regressors is similar in
spirit to the Kinect body pose estimation work of \cite{Shotton13},
although the application is quite different.

\cite{Wu15} recently investigated the learning of simple kinematics,
in the context of objects sliding down ramps. Similar to
\cite{Battaglia13}, they also used a
top-down 3D physics engine to map from a hypothesis of object mass,
shape, friction etc. to image space. Inference relies on MCMC,
initialized to the output of convnet-based estimates of the
attributes. As in our work, their
evaluations are performed on real data and the model predictions
correlate reasonably with human judgement.

Prior work in reinforcement learning has used synthetic data from games
to train bottom-up models. In particular, \cite{Mnih15} and \cite{Lillicrap15}
trained deep convolutional networks with reinforcement learning directly on image pixels from simulations
to learn policies for Atari games and the TORCS driving simulator, respectively.
% A policy network for a driving simulator may encode certain physical concepts such
% as depth and perspective, although this was not examined.

A number of works in cognitive science have explored intuitive
physics, for example, in the context of liquid dynamics \cite{Bates15}, ballistic
motion \cite{Smith13} and gears and pulleys \cite{Hegarty04}. The latter
finds that people perform ``mental simulation'' to answer questions about gears, pulleys,
etc., but some form of implicit bottom-up reasoning is involved too.

%The literature also contains some debate about the limitations of
%simulation and probabilistic modeling, e.g.~\cite{davis2016scope} and \cite{goodman2015relevant}.
%Implicit reasoning would seem a viable alternative, although the
%cognitive process likely has little similarity to our convnet models.
%Origin of Concepts \cite{Carey09}: \rob{Try to work in somewhere}.

In computer vision, a number of works have used physical reasoning to
aid scene understanding
\cite{zheng2015scene,koppula2016anticipating}. For example,
\cite{Jia15} fit cuboids to RGBD data and use their centroids to
search for scene interpretations that are statically stable.

% Mention  \cite{Silberman12}??

\section{Methods}

\subsection{UETorch}

UETorch is a package that embeds the Lua/Torch machine learning
environment directly into the UE4 game loop, allowing for fine-grained
scripting and online control of UE4 simulations through Torch. Torch
is well-suited for game engine integration because Lua is the dominant
scripting language for games, and many games including UE4 support Lua
scripting.  UETorch adds additional interfaces to capture
screenshots, segmentation masks, optical flow data, and control of the
game through user input or direct modification of game state. Since
Torch runs inside the UE4 process, new capabilities can be easily
added through FFI without defining additional interfaces/protocols for
inter-process communication. UETorch simulations can be run faster
than real time, aiding large-scale training.  The \texttt{UETorch} package can be downloaded
freely at \url{http://github.com/facebook/UETorch}.

\subsection{Data Collection}

\textbf{Synthetic}

\begin{figure}
\centerline{\includegraphics[width=\columnwidth]{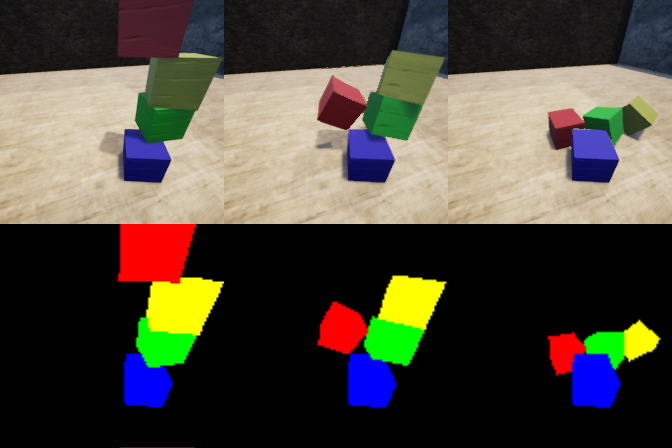}}
\caption{Recorded screenshots and masks at 1-second intervals from the Unreal Engine block simulation.}
\label{fig:synth_data_gen}
\end{figure}

A simulation was developed in UETorch that generated vertical stacks of
2, 3, or 4 colored blocks in random configurations. The block position and orientation, camera
position, background textures, and lighting were randomized at each
trial to improve the transferability of learned features.
In each simulation, we recorded the outcome (did it fall?) and captured screenshots and
segmentation masks at 8 frames/sec. Frames and masks from a representative 4-block simulation
are shown in \fig{synth_data_gen}. A total of 180,000 simulations were performed, balanced across number of blocks and
stable/unstable configurations. 12,288 examples were reserved for testing.

\textbf{Real}

Four wooden cubes were fabricated and spray painted
red, green, blue and yellow respectively.
Manufacturing imperfections added a certain level of randomness to the stability
of the real stacked blocks, and we did not attempt
to match the physical properties of the real and synthetic blocks.
The blocks were manually stacked in configurations 2, 3 and 4 high
against a white bedsheet. A tripod mounted DSLR camera was used to film the
blocks falling at 60 frames/sec. A white pole was held against the top block
in each example, and was then rapidly lifted upwards, allowing unstable stacks
to fall (the stick
can be see in \fig{teaser}, blurred due to its rapid motion). Note
that this was performed even for stable configurations, to avoid
bias. Motion
of the blocks was only noticeable by the time the stick was several
inches away from top block. 493 examples were captured, balanced between stable/unstable configurations. The
totals for 2, 3 and 4 block towers were 115, 139 and 239 examples
respectively.

\subsection{Human Subject Methodology}
\begin{figure}
\centerline{
\includegraphics[width=1\columnwidth]{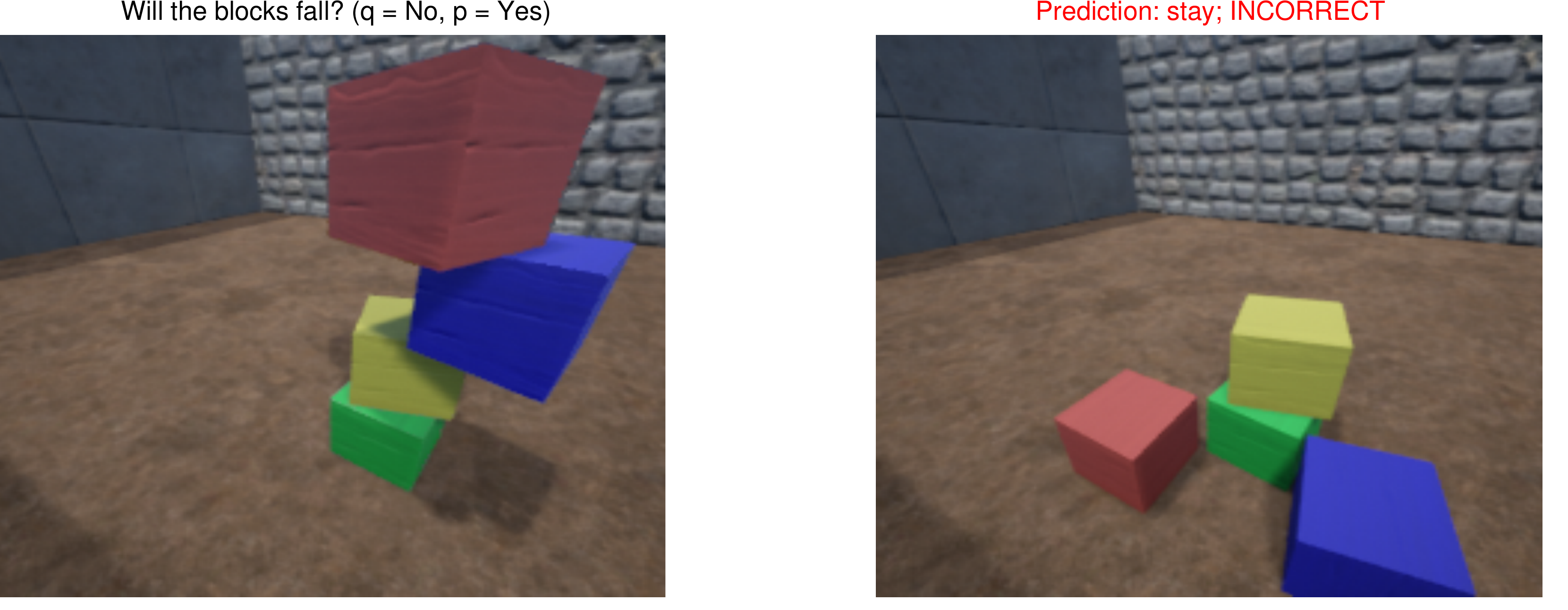}
}
\caption{The interface used for human experiments. At each turn, the
  subject is shown an image on the left and tries to predict if the stack
  will fall or not. No time limit is imposed. During training phase, the
  subject receives feedback on their prediction, by showing them the
  outcome image on the right. }
\label{fig:gui}
\end{figure}

To better understand the challenge posed about our datasets, real and
synthetic, we asked 10 human subjects to evaluate the images in a
controlled experiment. Participants were asked to give a binary prediction
regarding the outcome of the blocks (i.e. falling or not). During the training
phase, consisting of 50 randomly drawn examples, participants were shown the final
frame of each example, along with feedback as to whether their
choice was correct or not (see \fig{gui}). Subsequently, they were tested using
100 randomly drawn examples (disjoint from the training set). During
the test phase, no feedback was provided to the individuals regarding the
correctness of their responses.

\subsection{Model Architectures}
\label{sec:models}
We trained several convolutional network (CNN) architectures on the synthetic blocks dataset.
We trained some architectures on the binary fall prediction task only,
and others on jointly on the fall prediction and mask prediction tasks.

\textbf{Fall Prediction}

\label{sec:model-fall}
We trained the ResNet-34 \cite{He15} and Googlenet \cite{Szegedy14}
networks on the fall prediction task. These models were pre-trained on
the Imagenet dataset \cite{ILSVRC15}. We replaced the final linear layer
with a single logistic output and fine-tuned the entire network with
SGD on the blocks dataset. Grid search was performed over learning rates.

% We additionally trained Googlenet from scratch on the fall prediction task, to
% examine whether pre-training on Imagenet improves the prediction accuracy on the
% synthetic dataset or its transferability to real block images.

\textbf{Fall+Mask Prediction}

\begin{figure*}
\centerline{\includegraphics[width=2\columnwidth]{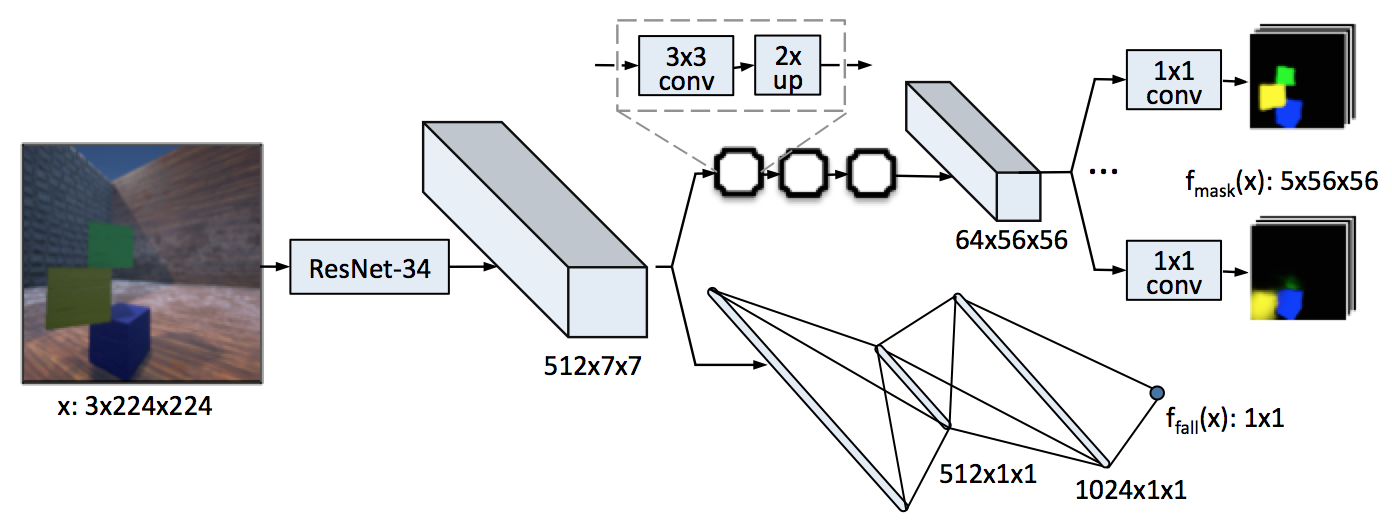}}
\caption{Architecture of the PhysNet network.}
\label{fig:PhysNet}
\end{figure*}

We used deep mask networks to predict the segmentation trajectory of falling
blocks at multiple future times (0s,1s,2s,4s) based on an input image. Each mask pixel is a multi-class
classification across a background class and four foreground (block color) classes. A fall
prediction is also computed.

DeepMask \cite{Pinheiro15} is an existing mask prediction network trained for instance segmentation,
and has the appropriate architecture for our purposes.
We replaced the binary mask head with a multi-class SoftMax, and replicated this $N$ times
for mask prediction at multiple points in time.

We also designed our own mask prediction network, PhysNet, that was suited to mask \textit{prediction}
rather than just segmentation. For block masks, we desired (i) spatially local
and translation-invariant (i.e. convolutional) upsampling from coarse image features to masks,
and (ii) more network depth at the coarsest spatial resolution, so the network could reason about
block movement. Therefore, PhysNet take the $7\times 7$ outputs from ResNet-34,
and performs alternating upsampling and convolution to arrive at $56\times 56$ masks.
The PhysNet architecture is shown in \fig{PhysNet}. We use the Resnet-34 trunk in PhysNet
for historical reasons, but our experiments show comparable results with a Googlenet trunk.

% The training loss for all mask networks is

% \begin{equation}
% \mathcal{L}(\textbf{w}) = \frac{1}{N} \sum_{n=1}^N{
%   \left[
%        H_2(f_n,p_n) +
%        c_m \sum_{x,y=1}^{56} {H_5(m_{nxy},q_{nxy})}
%   \right]
% }
% \end{equation}

% where $N$ is the number of training examples, $H_c(\cdot,\cdot)$ is the cross-entropy loss over $c$ classes,
% $f_n$ and $p_n=p_n(\bf{w})$ are the fall label and prediction of training example $n$, and $m_{nxy}$ and $q_{nxy}=q_{nxy}(\bf{w})$ are the
% mask label and prediction of training example $n$ at location $(x,y)$. $c_m$ is a hyperparameter controlling the
% weight of fall vs. mask training.

The training loss for mask networks is the sum of a binary cross-entropy loss for fall prediction and
a pixelwise multi-class cross-entropy loss for each mask. A hyperparameter controls the relative weight of
these losses.

\textbf{Baselines}
\label{sec:baselines}
As a baseline, we perform logistic regression either directly on image pixels, or on pretrained Googlenet features,
to predict fall and masks. To reduce the number of parameters, the pixels-to-mask matrix is factored with an intermediate dimension 128.
For fall prediction, we also try $k$-Nearest-Neighbors ($k=10$) using Googlenet last-layer image features.

\vspace{-3mm}
\subsection{Evaluation}
\label{sec:eval}
We compare fall prediction accuracy on synthetic and real images, both
between models and also between model and human performance. We also train
models with a held-out block tower size and test them on the held out tower size,
to evaluate the transfer learning capability of these models models to different block tower sizes.

We evaluate mask predictions with two criteria: mean mask IoU and log
likelihood per pixel. We define mean mask IoU as the intersection-over-union of the mask label with the binarized prediction
for the $t=4s$ mask, averaged over each foreground class present in the mask label.
\begin{equation}
\label{eq:iou}
MIoU(\textbf{m},\textbf{q}) = \frac{1}{N} \sum_{n=1}^{N}
{\left[
  \frac{1}{|C_n|}
  \sum_{c\in C_n}
    { IoU(m_{nc},\hat{q}_{nc}) }
\right]
}
\end{equation}
where $m_{nc}$ is the set of pixels of class $c$ in mask $n$,
$C_n = \{ c : c\in \{1,2,3,4\} \wedge |m_{nc}| > 0 \}$
is the set of foreground classes present in mask $n$,
$\hat{q}_{nc}$ is the set of pixels in model output $n$ for which $c$ is the highest-scoring class,
and $IoU(m_1,m_2) = \frac {|m_1\cap m_2 |} {|m_1\cup m_2 |}$.

The mask IoU metric is intuitive but problematic because it uses binarized masks. For example, if the model
predicts a mask with 40\% probability in a region, the Mask IoU for that block will be 0 whether
or not the block fell in that region. The quality of the predicted mask confidences is better
captured by log likelihood.

The log likelihood per pixel is defined as the log likelihood of the correct final mask
under the predicted (SoftMax) distribution, divided by the number of pixels. This is
essentially the negative mask training loss.

% \begin{equation}
% \label{eq:log-likelihood}
% LL(\textbf{m},\textbf{q}) = \frac{1}{N*56*56} \sum_{n=1}{N} { \sum_{x,y=1}^{56}{ \sum_{c=0}^{4} { m_{nxyc} \log q_{nxyc} } } }
% \end{equation}

Since the real data has a small number of examples ($N=493$ across all blocks sizes),
we report an estimated confidence interval for the model prediction on real examples.
We estimate this interval as the standard deviation of a binomial distribution with $p$
approximated by the observed accuracy of the model.

\section{Results}

% \begin{table*}[t]
% \begin{center}
% %\begin{tabular}{| l | c | c | c | c |}
% \begin{tabular}{ l  c  c  c  c }
% \toprule
% \bf{Model} & \bf{Fall Accuracy (\%)} & \bf{Fall Accuracy (\%)} & \bf{Mean Mask IoU (\%)} & \bf{Mask Loss/px} \\
%                  & \bf{(synthetic)} & \bf{(real)}              & \bf{(synthetic)}        & \bf{(synthetic)}\\
% \midrule
% VGG-A         & 83.8 & 60.3 & & \\
% Googlenet     & 86.5 & \textbf{69.8} & & \\
% ResNet        & 84.7 & 67.8 & & \\
% Deepmask-A    & 83.5 & 66.4 & 31.7 & 0.452 \\
% Deepmask-B    & 49.0 & 47.0 & 42.4 & 0.299 \\
% VGGConv       & 85.4 & 67.0 & 72.7 & 0.139 \\
% PhysNet    & \textbf{89.1} & 67.2 & \textbf{75.4} & \textbf{0.107} \\
% \hline
% \hline
% \bf{Baseline} & & & & \\
% Logistic Reg           & 52.9 & 49.3 & 30.0 & 0.636 \\
% VGG-A Log. Reg.        & 64.1 & 61.3 & 23.4 & 0.467 \\
% Googlenet Log. Reg.    & 65.8 & 63.1 & 23.4 & 0.536 \\
% VGG-A kNN              & 54.8 & 50.9 &      &       \\
% Googlenet kNN          & 59.6 & 50.9 &      &       \\
% & & & & \\
% \hline
% \bottomrule
% \end{tabular}
% \caption{Models lorem ipsum.}
% \label{tab:models}
% \end{center}
% \end{table*}

\subsection{Fall Prediction Results}

\tab{models_fall} compares the accuracy for fall prediction of several deep networks and baselines
described in \secc{models}. Convolutional networks perform well at fall prediction, whether trained
in isolation or jointly with mask prediction. The best accuracy on synthetic data is achieved with PhysNet,
which is jointly trained on masks and fall prediction. Accuracy on real data for all convnets is
within their standard deviation.

As an ablation study, we also measured the performance of Googlenet without Imagenet pretraining.
Interestingly, while the model performed equally well on synthetic data with and without pretraining,
only the pretrained model generalized well to real images (\tab{models_fall}).

\begin{table}
\begin{center}
\small
%\begin{tabular}{| l | c | c | c | c |}
\begin{tabular}{ l  c  c }
\toprule
\bf{Model} & \bf{Fall Acc. (\%)} & \bf{Fall Acc. (\%)} \\
           & \bf{(synthetic)}    & \bf{(real)}              \\
\midrule
\bf{Baselines} & & \\
Random                 & 50.0 & 50.0 {$\scriptstyle \pm$} \scriptsize 2.2 \\
Pixel Log. Reg         & 52.9 & 49.3 {$\scriptstyle \pm$} \scriptsize 2.2 \\
Googlenet Log. Reg.    & 65.8 & 62.5 {$\scriptstyle \pm$} \scriptsize 2.2 \\
Googlenet kNN          & 59.6 & 50.9 {$\scriptstyle \pm$} \scriptsize 2.2 \\
\hline
\bf{Classification Models} & & \\
ResNet-34              & 84.7          &         67.1  {$\scriptstyle \pm$} \scriptsize 2.1 \\
Googlenet              & 86.5          & \textbf{69.0} {$\scriptstyle \pm$} \scriptsize 2.1 \\
Googlenet              & 86.5          &         59.2 {$\scriptstyle \pm$} \scriptsize 2.2  \\
(no pretraining)       &               &                                                    \\
\hline
\bf{Mask Prediction Models} & & \\
DeepMask               & 83.5          &         66.1  {$\scriptstyle \pm$} \scriptsize 2.1 \\
PhysNet                & \textbf{89.1} &         66.7  {$\scriptstyle \pm$} \scriptsize 2.1 \\
\bottomrule
\end{tabular}
\caption{Fall prediction accuracy of convolutional networks on synthetic and real data.
The models substantially outperform baselines, and all have similar performance whether
trained singly or jointly with the mask prediction task.
Training Googlenet without Imagenet pretraining does not affect performance on synthetic examples,
but degrades generalization to real examples.
Baselines are described in \secc{baselines}.
}
\label{tab:models_fall}
\end{center}
\end{table}

% \begin{table}
% \begin{center}
% \small
% %\begin{tabular}{| l | c | c | }
% \begin{tabular}{ l  c  c  }
% \toprule
% \bf{Pretraining} & \bf{Fall Acc. (\%)} & \bf{Fall Acc. (\%)} \\
%                  & \bf{(synthetic)} & \bf{(real)} \\
% \midrule
% None     & 86.5 & 59.2 {$\scriptstyle \pm$} \scriptsize 2.2 \\
% Imagenet & 86.5 & 69.0 {$\scriptstyle \pm$} \scriptsize 2.1 \\
% \bottomrule
% \end{tabular}
% \caption{Googlenet fall prediction performance on real and synthetic data,
% with and without Imagenet pretraining. The models perform equally well on
% synthetic data, but the pretrained model generalizes substantially better to
% real images.}
% \label{tab:googlenet_pretrain}
% \end{center}
% \end{table}

\textbf{Occlusion Experiments}

\begin{figure}
\vskip -0.03in
\begin{tabular}{p{0.45\columnwidth} p{0.45\columnwidth}}
        A & B
\end{tabular}
\vskip -0.03in
\centerline{
\includegraphics[width=0.50\columnwidth]{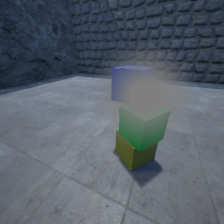}
\includegraphics[width=0.50\columnwidth]{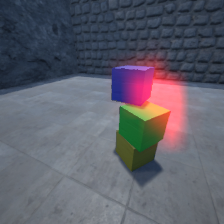}
}
\vskip -0.03in
\begin{tabular}{p{0.45\columnwidth} p{0.45\columnwidth}}
        C & D
\end{tabular}
\vskip -0.03in
\centerline{
\includegraphics[width=0.50\columnwidth]{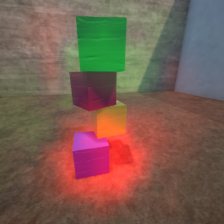}
\includegraphics[width=0.50\columnwidth]{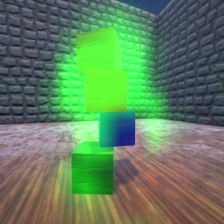}
}
\caption{
\textbf{A}: Example of Gaussian occlusion mask, applied in a sliding window to generate
fall prediction heatmaps.
\textbf{B--D}: Heatmaps of predictions from occluded images. A \textit{green} overlay means that an occlusion in this region \textit{increases} the predicted probability of falling,
while a \textit{red} overlay means the occlusion \textit{decreases} the predicted probability of falling.
The model focuses on unstable interfaces (\textbf{B,C}), or stabilizing blocks that prevent the tower
from falling (\textbf{D}).
}
\label{fig:blur}
\end{figure}

We performed occlusion experiments to determine which regions of the
block images affected the models' fall predictions. A Gaussian patch
of gray pixels with standard deviation $20\%$ of the image width was
superimposed on the image in a
$14\times 14$ sliding window to occlude parts of the image, as shown
in \fig{blur}A. The PhysNet model was evaluated on each occluded
image, and the difference in the fall probability predicted from the
baseline and occluded images were used to produce heatmaps, shown in
\fig{blur}B-D. These figures suggest that the model makes its
prediction based on relevant local image features rather than
memorizing the particular scene. For example, in \fig{blur}B, the
model prediction is only affected by the unstable interface between
the middle and top blocks.

\textbf{Model vs. Human Performance}

\begin{figure}
\begin{center}
\centerline{
\includegraphics[width=0.42\columnwidth]{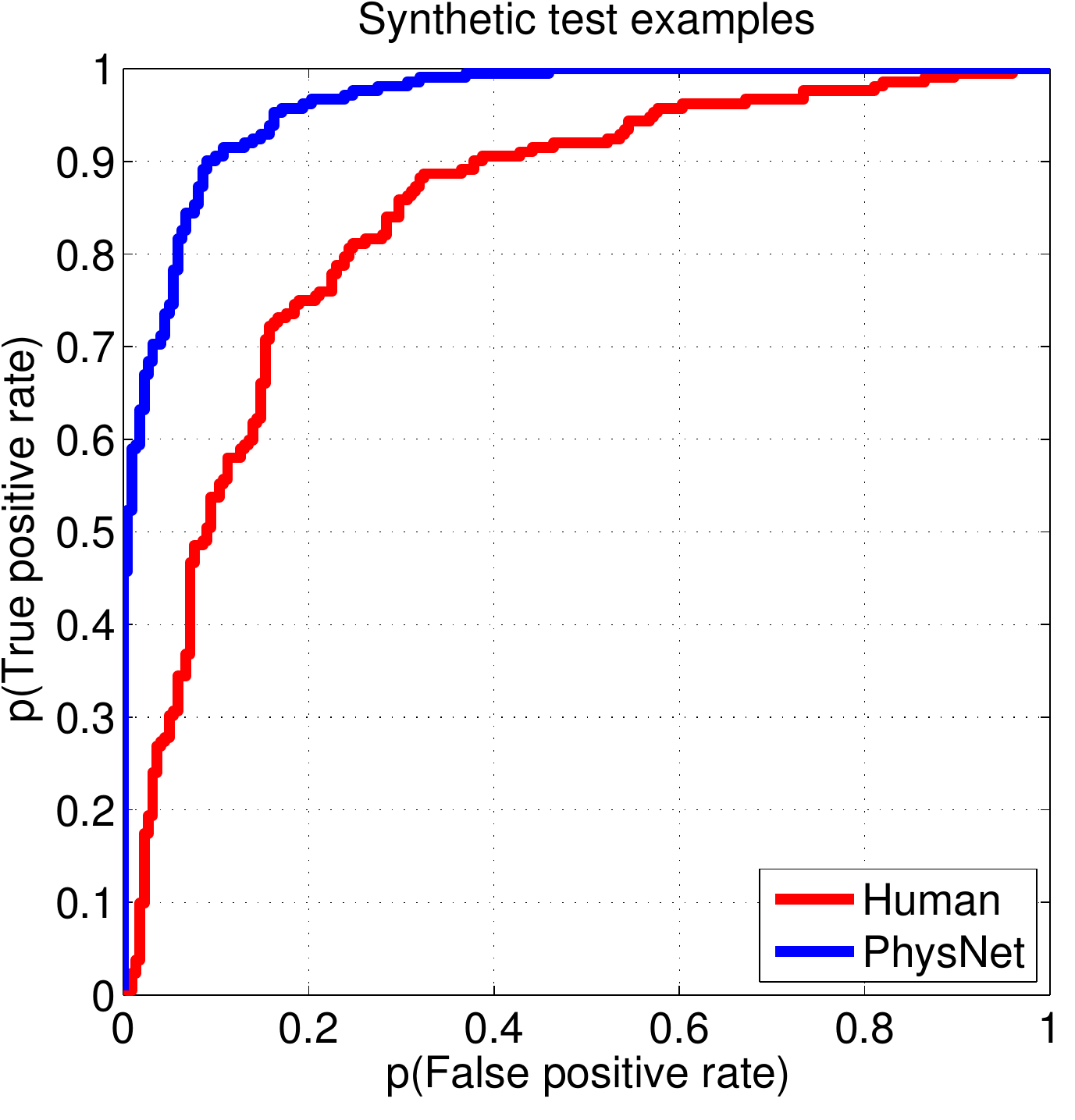}
\includegraphics[width=0.55\columnwidth]{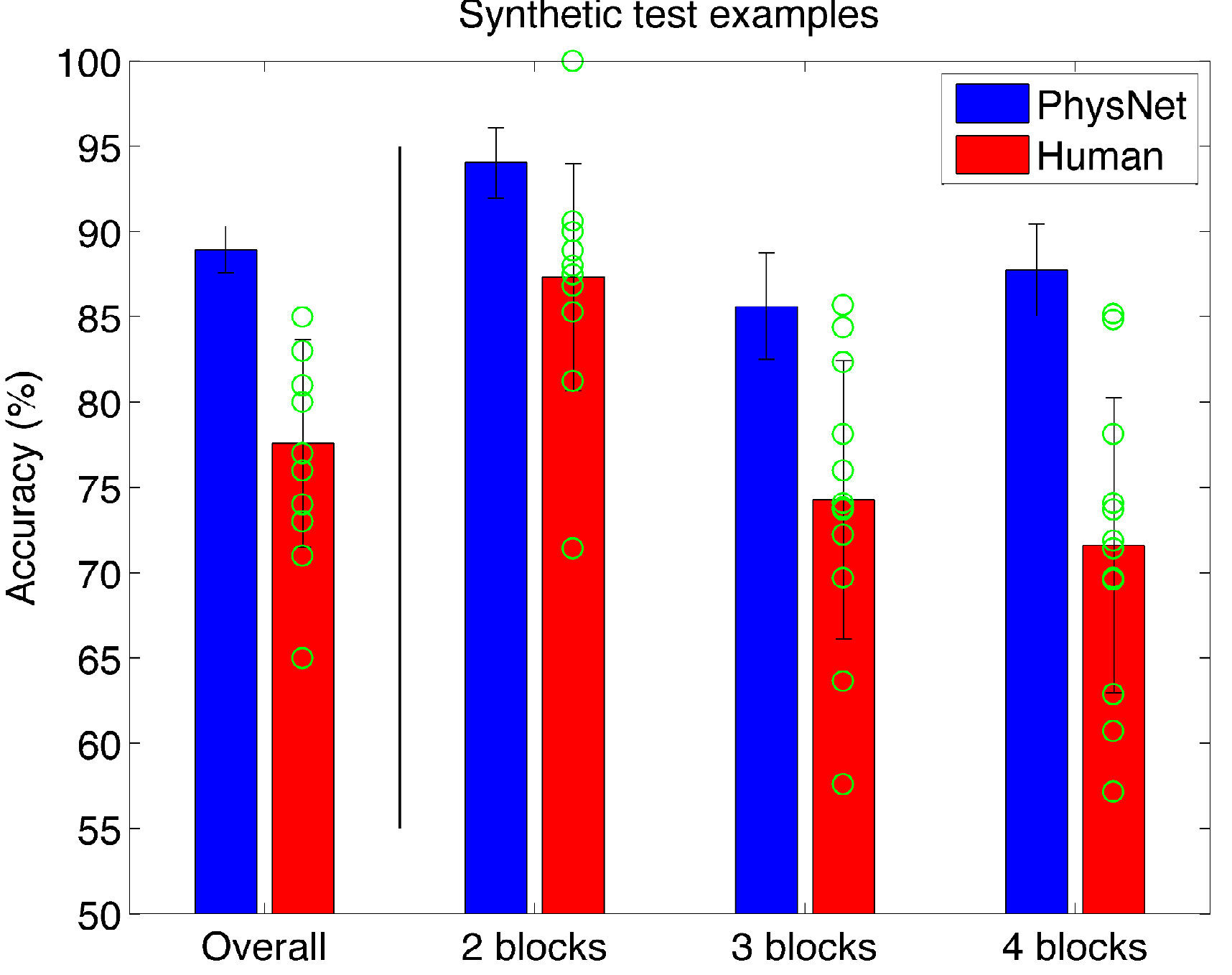}
}
\centerline{
\includegraphics[width=0.42\columnwidth]{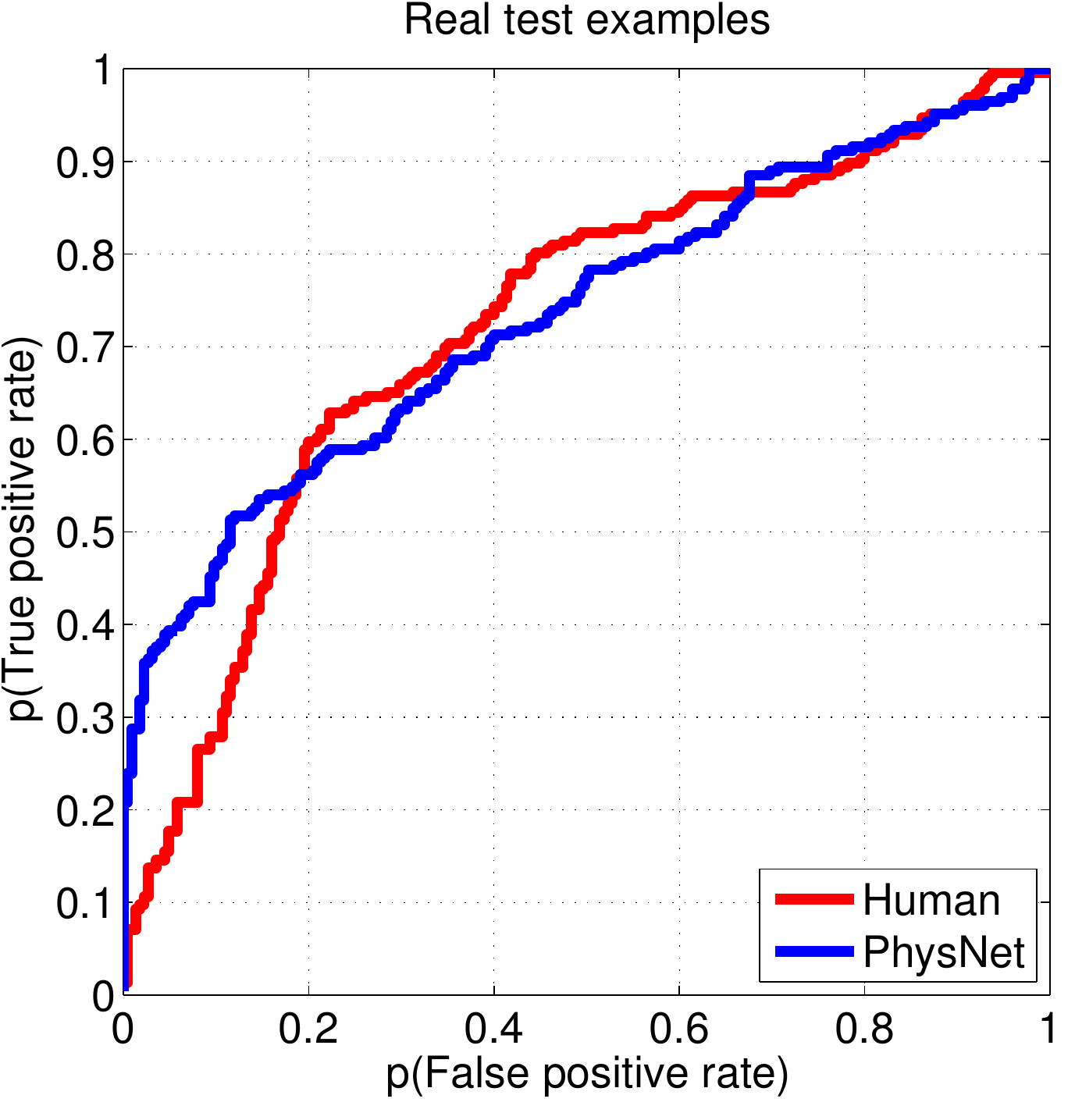}
\includegraphics[width=0.55\columnwidth]{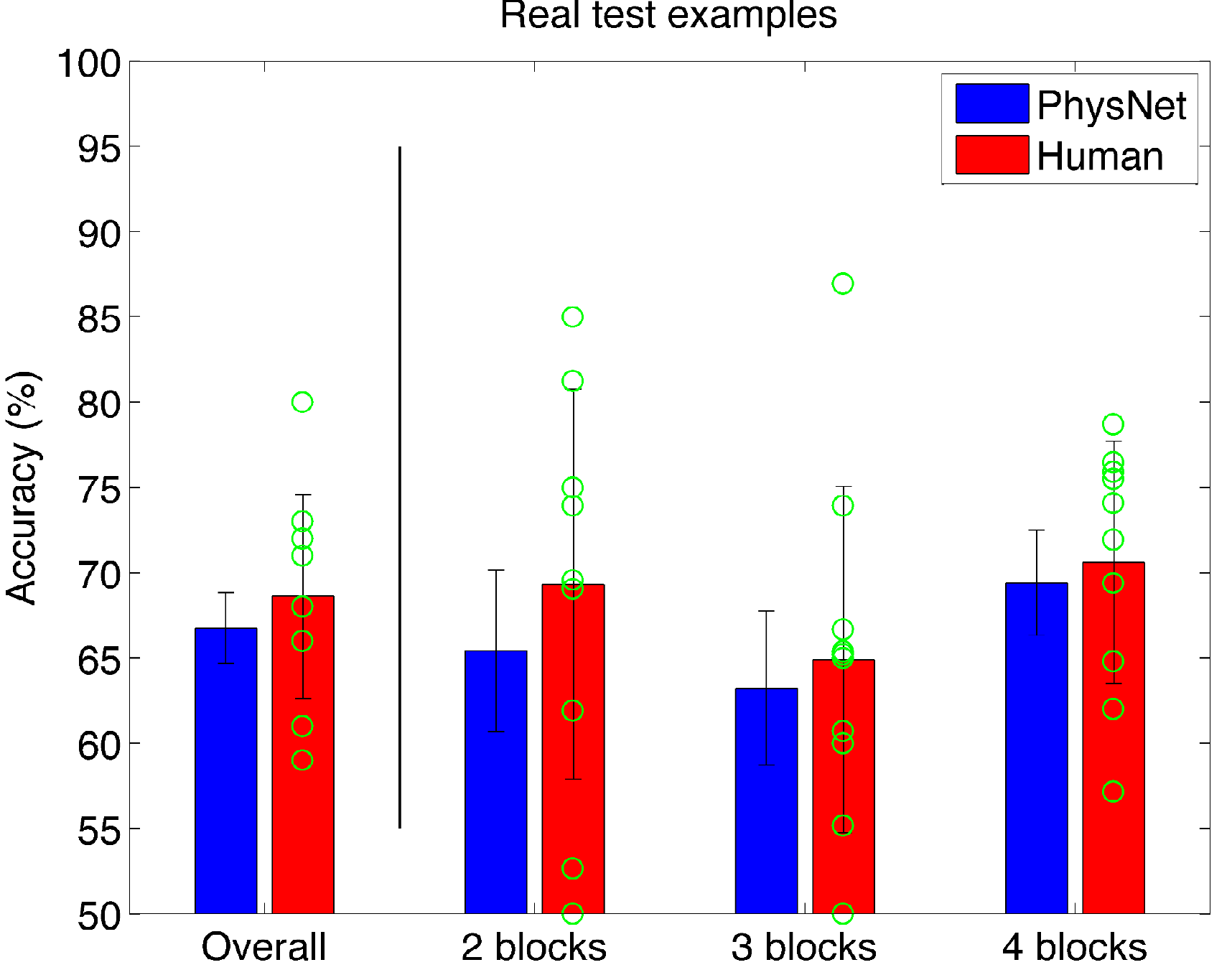}
}
\caption{Plots comparing PhysNet accuracy to human performance
  on real (Top) and synthetic (Bottom) test examples.
  Left: ROC plot comparing human and model predictions.
  Right: a breakdown of the performance for differing numbers of
  blocks. For humans, the mean performance is shown, along with
  the performance of individual subjects (green circles).
  Overall, the PhysNet model is better
  than even the best performing of the human subjects on synthetic data.
  On real data, PhysNet performs similarly to humans.}
\label{fig:results_human}
\end{center}
\end{figure}

\fig{results_human} compares PhysNet to 10 human
subjects on the same set of synthetic and real test images. ROC curves comparing
human and model performance are generated by using the fraction of test
subjects predicting a fall as a proxy for confidence, and comparing this
to model confidences.

Overall, the model
convincingly outperforms the human subjects on synthetic data, and is
comparable on real data. Interestingly, the
correlation between human and model confidences on both real and synthetic data
($\rho=(0.69,0.45)$) is higher
than between human confidence and ground truth ($\rho=(0.60,0.41)$),
showing that our model agrees quite closely with human judgement.

% -----------------------------------------

\begin{figure*}[p]
\begin{center}
\centerline{
\begin{tabular}{p{0.00\columnwidth}}\textbf{A}\end{tabular} \includegraphics[width=0.95\columnwidth]{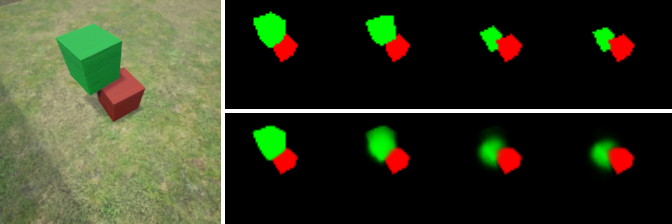}
\begin{tabular}{p{0.00\columnwidth}}\textbf{G}\end{tabular} \includegraphics[width=0.95\columnwidth]{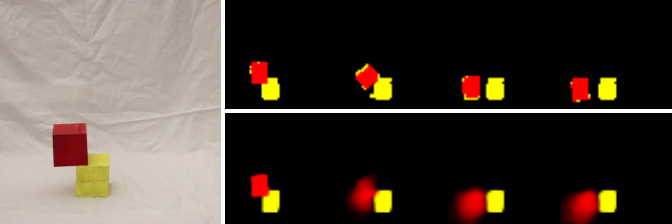}
} \vskip 0.05in
\centerline{
\begin{tabular}{p{0.00\columnwidth}}\textbf{B}\end{tabular} \includegraphics[width=0.95\columnwidth]{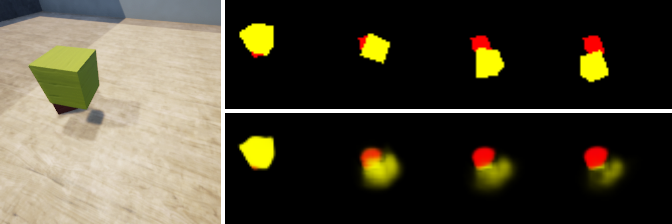}
\begin{tabular}{p{0.00\columnwidth}}\textbf{H}\end{tabular} \includegraphics[width=0.95\columnwidth]{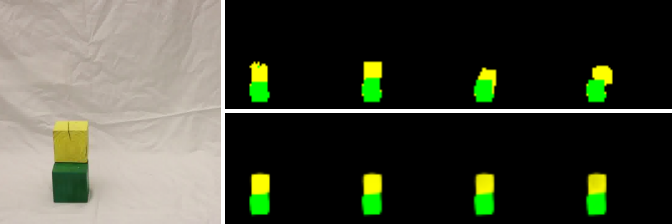}
} \vskip 0.05in
\centerline{
\begin{tabular}{p{0.00\columnwidth}}\textbf{C}\end{tabular} \includegraphics[width=0.95\columnwidth]{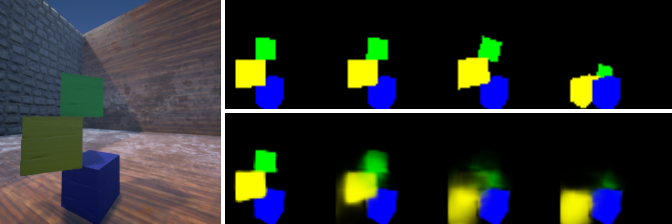}
\begin{tabular}{p{0.00\columnwidth}}\textbf{I}\end{tabular} \includegraphics[width=0.95\columnwidth]{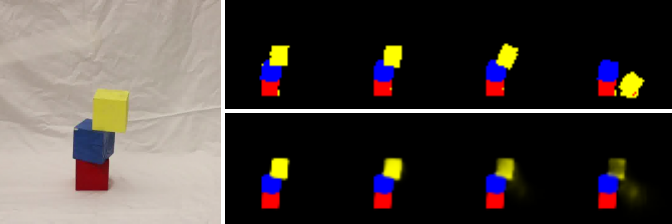}
} \vskip 0.05in
\centerline{
\begin{tabular}{p{0.00\columnwidth}}\textbf{D}\end{tabular} \includegraphics[width=0.95\columnwidth]{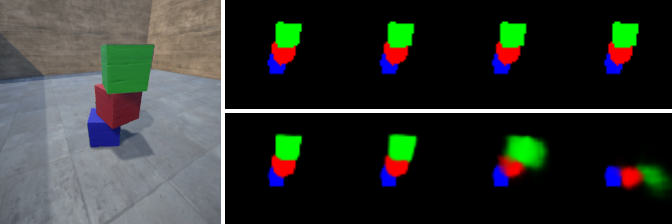}
\begin{tabular}{p{0.00\columnwidth}}\textbf{J}\end{tabular} \includegraphics[width=0.95\columnwidth]{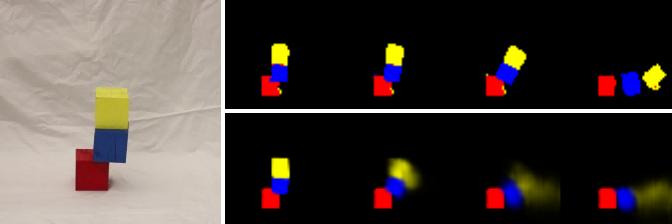}
} \vskip 0.05in
\centerline{
\begin{tabular}{p{0.00\columnwidth}}\textbf{E}\end{tabular} \includegraphics[width=0.95\columnwidth]{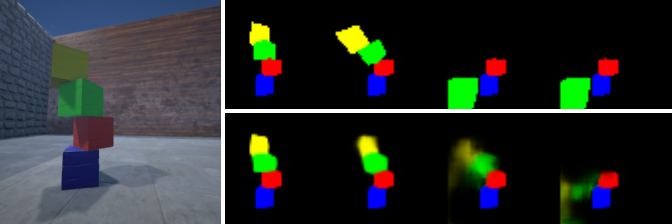}
\begin{tabular}{p{0.00\columnwidth}}\textbf{K}\end{tabular} \includegraphics[width=0.95\columnwidth]{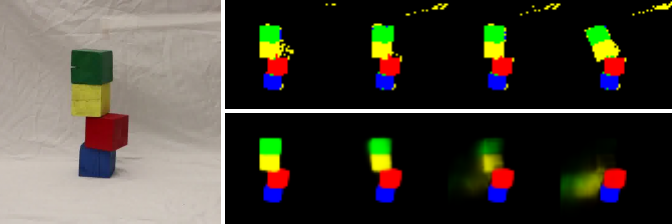}
} \vskip 0.05in
\centerline{
\begin{tabular}{p{0.00\columnwidth}}\textbf{F}\end{tabular} \includegraphics[width=0.95\columnwidth]{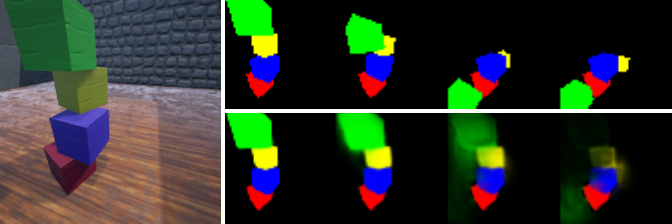}
\begin{tabular}{p{0.00\columnwidth}}\textbf{L}\end{tabular} \includegraphics[width=0.95\columnwidth]{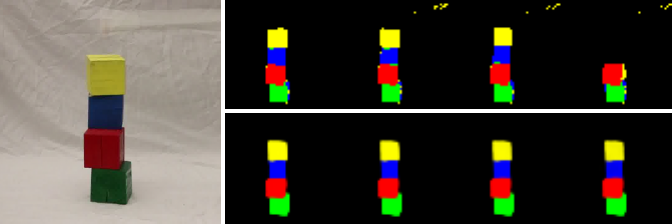}
}
\caption{
PhysNet mask predictions for synthetic (\textbf{A--F}) and real (\textbf{G--L}) towers of 2, 3, and 4 blocks.
The image at the left of each example is the initial frame shown to the model.
The top row of masks are the ground truth masks from simulation, at 0, 1, 2, and 4 seconds.
The bottom row are the model predictions, with the color intensity representing the predicted probability.
PhysNet correctly predicts fall direction and occlusion patterns for most synthetic examples,
while on real examples, PhysNet overestimates stability (\textbf{H,L}).
In difficult cases, Physnet produces diffuse masks due to uncertainty (\textbf{D--F,I}).
\textbf{B} is particularly notable, as PhysNet predicts the red block location from the
small patch visible in the initial image.
}
\label{fig:masks}
\end{center}
\end{figure*}

% \begin{figure}[p]
% \begin{center}

% \centerline{ \begin{tabular}{p{0.00\columnwidth}}\textbf{A}\end{tabular} \includegraphics[width=0.95\columnwidth]{real_blocks_169.png} } \vskip 0.05in
% \centerline{ \begin{tabular}{p{0.00\columnwidth}}\textbf{B}\end{tabular} \includegraphics[width=0.95\columnwidth]{real_blocks_181.png} } \vskip 0.05in
% \centerline{ \begin{tabular}{p{0.00\columnwidth}}\textbf{C}\end{tabular} \includegraphics[width=0.95\columnwidth]{real_blocks_198.png} } \vskip 0.05in
% \centerline{ \begin{tabular}{p{0.00\columnwidth}}\textbf{D}\end{tabular} \includegraphics[width=0.95\columnwidth]{real_blocks_202.png} } \vskip 0.05in
% \centerline{ \begin{tabular}{p{0.00\columnwidth}}\textbf{E}\end{tabular} \includegraphics[width=0.95\columnwidth]{real_blocks_274.png} } \vskip 0.05in
% \centerline{ \begin{tabular}{p{0.00\columnwidth}}\textbf{F}\end{tabular} \includegraphics[width=0.95\columnwidth]{real_blocks_287.png} }
% \caption{
% PhysNet mask predictions for real block towers, in the same format as \fig{masks_synth}.
% Mask predictions on real towers are less accurate than on synthetic towers, and tend to predict
% that moderately unstable towers are stable (\textbf{B,F}).}
% %\vskip 0.75in
% \label{fig:masks_real}
% \end{center}
% \end{figure}

\subsection{Mask Prediction Results}

\tab{models_mask} compares mask prediction accuracy of the DeepMask and PhysNet networks
described in \secc{models}. PhysNet achieves the best performance on both Mean Mask IoU and
Log Likelihood per pixel (see \secc{eval}), substantially outperforming DeepMask and baselines.
Predicting the mask as equal to the initial ($t=0$) mask has a high Mask IoU due to the deficiencies
in that metric described in \secc{eval}.

Examples of PhysNet mask outputs on synthetic and real
data are shown in \fig{masks}. We only show masks for examples that are
predicted to fall, because predicting masks for stable towers is easy and the outputs are typically perfect.
The mask outputs from PhysNet are typically quite reasonable for falling 2- and 3-block synthetic towers, but have more
errors and uncertainty on 4-block synthetic towers and most real examples. In these cases, the masks are often
highly diffuse, showing high uncertainty about the trajectory. On real examples, model predictions and masks
are also skewed overstable, likely because of different physical properties of the real and simulated blocks.

\begin{table}
\begin{center}
\small
%\begin{tabular}{| l | c | c | c | c |}
\begin{tabular}{ l  c  c }
\toprule
\bf{Model} & \bf{Mask IoU (\%)} & \bf{Log Likelihood/px} \\
           & \bf{(synthetic)}   & \bf{(synthetic)}\\
\midrule

DeepMask               & 42.4 & -0.299 \\
PhysNet             & \textbf{75.4} & \textbf{-0.107} \\
\hline
\hline
\bf{Baseline} & &\\
Pixel Log. Reg.        & 29.6 & -0.562 \\
Googlenet Log. Reg.    & 23.8 & -0.492 \\
Mask @ $t=0$           & 72.0 & $-\infty$ \\
Class-Constant         & 0    & -0.490  \\
\bottomrule
\end{tabular}
\caption{Mask prediction accuracy of DeepMask and our PhysNet network.
The metrics used are described in \secc{eval}; baselines are described in \secc{baselines}.
As an additional IoU baseline we evaluate the $t=0$ mask as a prediction of the final mask,
and as a log likelihood baseline we predict each pixel as the average likelihood of that class
in the data.
The PhysNet network provides the highest accuracy in both metrics. Mask examples are shown in \fig{masks}.
}
\label{tab:models_mask}
\end{center}
\end{table}

\begin{figure}
\centerline{ \begin{tabular}{p{0.00\columnwidth}}\textbf{A}\end{tabular} \includegraphics[width=0.95\columnwidth]{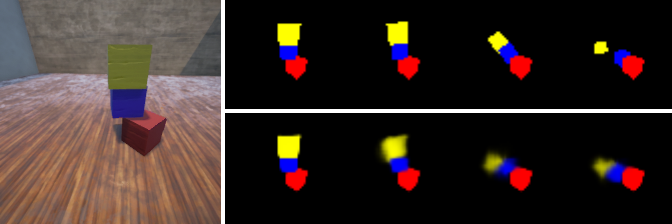} } \vskip 0.05in
\centerline{ \begin{tabular}{p{0.00\columnwidth}}\textbf{B}\end{tabular} \includegraphics[width=0.95\columnwidth]{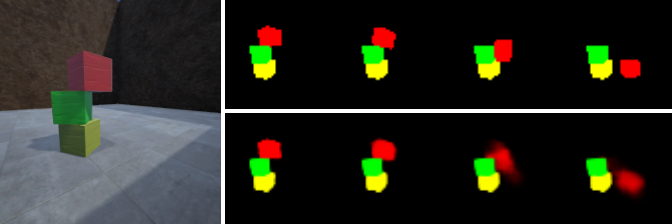} } \vskip 0.05in
\centerline{ \begin{tabular}{p{0.00\columnwidth}}\textbf{C}\end{tabular} \includegraphics[width=0.95\columnwidth]{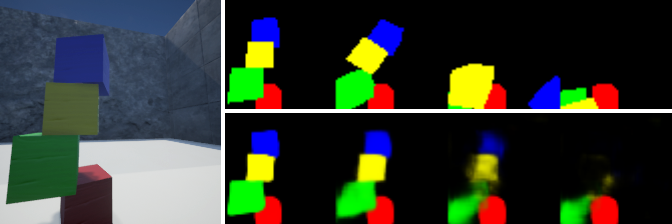} } \vskip 0.05in
\centerline{ \begin{tabular}{p{0.00\columnwidth}}\textbf{D}\end{tabular} \includegraphics[width=0.95\columnwidth]{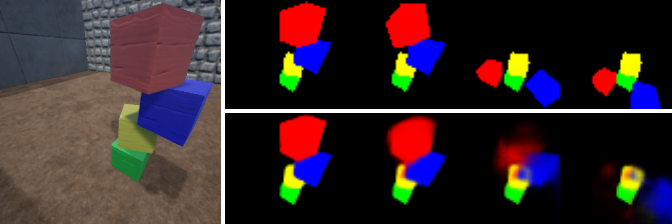} }
\caption{
PhysNet mask predictions on a tower size (3 or 4 blocks) that the network was \textit{not}
trained on. Mask predictions for 3 blocks (\textbf{A--B}) still capture the dynamics well
even though the network never saw towers of 3 blocks. Mask predictions for 4 blocks capture some
of the dynamics but show some degradation.
}
\label{fig:masks_transfer}
\end{figure}

\subsection{Evaluation on Held-Out Number of Blocks}

\tab{block_transfer} compares the performance of networks that had
either 3- or 4-block configurations excluded from the training set. While the accuracy of these
networks is lower on the untrained class relative to a fully-trained model, it's still relatively high -- comparable to
human performance. The predicted masks on the untrained number of blocks also continue to
capture the fall dynamics with reasonably accuracy. Some examples are shown in \fig{masks_transfer}.

\begin{table*}
\begin{center}
\small
%\begin{tabular}{ l | c | c | c | c | c | c | c }
\begin{tabular}{ l  l | c  c  c | c  c  c | c c c }
\toprule
\bf{Model} & \bf{\# Blocks} & \multicolumn{3}{c}{\bf{Accuracy (\%) (synth.)}} & \multicolumn{3}{c}{\bf{Accuracy (\%) (real)}} & \multicolumn{3}{c}{\bf{Mask Log Likelihood/px (synth.)}}\\
           & \bf{Training}  & \bf{2} & \bf{3} & \bf{4}                           & \bf{2} & \bf{3} & \bf{4}                      & \bf{2} & \bf{3} & \bf{4}                     \\
\midrule
Googlenet   & 2,3,4         & 92.6   &                     86.7  &                     82.3 &        69.6 {$\scriptstyle \pm$} \scriptsize 4.3 &                     69.8 {$\scriptstyle \pm$} \scriptsize 3.9 &                     69.9 {$\scriptstyle \pm$} \scriptsize 3.0 & & & \\
Googlenet   & 2,3           & 93.7   &                     85.9  & \cellcolor{blue!25} 71.3 &        65.2 {$\scriptstyle \pm$} \scriptsize 4.4 &                     66.9 {$\scriptstyle \pm$} \scriptsize 4.0 & \cellcolor{blue!25} 69.0 {$\scriptstyle \pm$} \scriptsize 3.0 & & & \\
Googlenet   & 2,4           & 93.3   & \cellcolor{blue!25} 82.0  &                     79.5 &        69.6 {$\scriptstyle \pm$} \scriptsize 4.3 & \cellcolor{blue!25} 66.9 {$\scriptstyle \pm$} \scriptsize 4.0 &                     70.7 {$\scriptstyle \pm$} \scriptsize 2.9 & & & \\
\hline
PhysNet  & 2,3,4            & 94.5   & 87.9                      &                     84.7 &        66.1 {$\scriptstyle \pm$} \scriptsize 4.4 &                     65.5 {$\scriptstyle \pm$} \scriptsize 4.0 &                     73.2 {$\scriptstyle \pm$} \scriptsize 2.9 & -0.035 & -0.096 & -0.190 \\
PhysNet  & 2,3              &  95.0  & 87.4                      & \cellcolor{blue!25} 77.3 &        60.0 {$\scriptstyle \pm$} \scriptsize 4.6 &                     64.0 {$\scriptstyle \pm$} \scriptsize 4.1 & \cellcolor{blue!25} 70.1 {$\scriptstyle \pm$} \scriptsize 2.9 & -0.042 & -0.125 & \cellcolor{blue!25} -0.362 \\
PhysNet  & 2,4              &  93.5  & \cellcolor{blue!25} 84.5  &                     83.6 &        55.7 {$\scriptstyle \pm$} \scriptsize 4.6 & \cellcolor{blue!25} 67.6 {$\scriptstyle \pm$} \scriptsize 4.0 &                     69.9 {$\scriptstyle \pm$} \scriptsize 3.0 & -0.040 & \cellcolor{blue!25} -0.154 & -0.268 \\
\bottomrule
\end{tabular}
\caption{Fall prediction accuracy for Googlenet and PhysNet trained on subsets of the block tower sizes,
and tested on the held-out block tower size (blue cells). Prediction accuracy on the held-out class is reduced,
but is still comparable to human performance (see \fig{results_human}). On real block data, performance
on the held out class is equivalent to the fully-trained model, to within standard deviation.
PhysNet mask predictions for held-out classes are only moderately degraded, and log likelihood scores are still superior to
DeepMask predictions (\tab{models_fall}). Physnet masks for the held-out class are shown in \fig{masks_transfer}.
 }
\label{tab:block_transfer}
\end{center}
\end{table*}

\section{Discussion}

Our results indicate that bottom-up deep CNN models can attain human-level performance at predicting how
towers of blocks will fall. We also find that these models' performance generalizes well to real images if
the models are pretrained on real data (\tab{models_fall}).

Several experiments provide evidence that the deep models we train are gaining knowledge about the dynamics of
the block towers, rather than simply memorizing a mapping from configurations to
outcomes. Most convincingly, the relatively small degradation in performance of the models on a tower size
that is not shown during training (\tab{block_transfer} \& \fig{masks_transfer}) demonstrates that the model must
be making its prediction based on local features rather than memorized exact block configurations. The occlusion experiments
in \fig{blur} also suggest that models focus on particular regions that confer stability or instability to a block configuration.
Finally, the poor performance of k-nearest-neighbors on Googlenet features in \tab{models_fall}
suggests that nearby configurations in Googlenet's pretrained feature space are not predictive of the stability of
a given configuration.

Compared to top-down, simulation-based models such as \cite{Battaglia13}, deep models require
far more training data -- many thousands of examples -- to achieve a high level of performance.
Deep models also have difficulty generalizing to examples far from their training data.
These difficulties arise because deep models must learn physics from scratch, whereas simulation-based models
start with strong priors encoded in the physics simulation engine.
Bottom-up and top-down approaches each have their advantages, and the precise combination of these systems in human reasoning
is the subject of debate (e.g.~\cite{davis2016scope} and \cite{goodman2015relevant}).
Our results suggest that deep models show promise for directly capturing common-sense
physical intuitions about the world that could lead to more powerful visual reasoning systems.

We believe that synthetic data from realistic physical simulations in UETorch are useful for
other machine learning experiments in vision, physics, and agent learning. The combination of synthetic
data and mask prediction constitutes a general framework for learning concepts such as object permanence,
3D extent, occlusion, containment, solidity, gravity, and collisions, that may be explored in the future.

% \adam{We have shown that the network has to learn something about dynamics rather than memorizing in several ways.
% Or at least it learns the right spatially local features to memorize.
% (i) Googlenet kNN, (ii) occlusion experiments, (iii) transfer masks to different number of blocks (most convincing)}
% \adam{Imagenet features + synth data is viable approach for learning real-world common-sense}
% \adam{Proposal: synthetic data + mask networks are a viable approach for learning a variety of common-sense
% visual reasoning tasks. Physics, occlusion, solidity, containment, etc. can all be modelled as predicting (segmentation or overlapping) masks}.
% \adam{mask prediction: how do you get realistic 'simulations' of the blocks falling rather than distributions?}

\section*{Acknowledgements}
The authors would like to thank: Soumith Chintala and Arthur Szlam for
early feedback on experimental design; Sainbayar Sukhbaatar for assistance
collecting the real-world block examples; Y-Lan Boureau for useful
advice regarding the human subject experiments; and Piotr Dollar for
feedback on the manuscript.

\clearpage
\bibliography{physnet}
\bibliographystyle{icml2016}

\end{document}